\documentclass[runningheads]{llncs}
\titlerunning{ACM for Intelligent CQA}
\authorrunning{M. M. Perera et al.} % Shortened author names for the header
\usepackage{array}
\usepackage{xcolor}
\usepackage{authblk}
\usepackage{longtable}
\usepackage{algorithm}
\usepackage{algorithmic}
\usepackage{amsmath}
\usepackage[T1]{fontenc}
\usepackage{amsmath}
\usepackage{graphicx}
\usepackage{float}
\usepackage{enumitem}
\usepackage{cite}
\usepackage{subcaption}
\usepackage{caption}
\usepackage[colorlinks=true, linkcolor=black, citecolor=black, urlcolor=blue]{hyperref} % Adjust colors here
\usepackage{cleveref}
\usepackage{multirow}
\crefname{figure}{Fig.}{Figs.}
\Crefname{figure}{Figure}{Figures}

\begin{document}

\title{Towards Adaptive Context Management for Intelligent Conversational Question Answering}

\author{Manoj Madushanka Perera\and Adnan Mahmood \and Kasun Eranda Wijethilake \and
Quan Z. Sheng}
\institute{School of Computing, Macquarie University, Sydney, NSW 2109, Australia
\email{manojmadushanka.perera@hdr.mq.edu.au},
\email{adnan.mahmood@mq.edu.au},
\email{kasuneranda.wijethilake@hdr.mq.edu.au},
\email{michael.sheng@mq.edu.au}}

\maketitle              % typeset the header of the contribution

\begin{abstract}
This particular paper introduces an \textit{Adaptive Context Management (ACM) framework} for the Conversational Question Answering (ConvQA) systems. The key objective of the ACM framework is to optimize the use of the conversation history by dynamically managing context for maximizing the relevant information provided to a ConvQA model within its token limit. Our approach incorporates a Context Manager (CM) Module, a Summarization (SM) Module, and an Entity Extraction (EE) Module in a bid to handle the conversation history efficaciously. The CM Module dynamically adjusts the context size, thereby preserving the most relevant and recent information within a model's token limit. The SM Module summarizes the older parts of the conversation history via a sliding window. When the summarization window exceeds its limit, the EE Module identifies and retains key entities from the oldest conversation turns. Experimental results demonstrate the effectiveness of our envisaged framework in generating accurate and contextually appropriate responses, thereby highlighting the potential of the ACM framework to enhance the robustness and scalability of the ConvQA systems.

\keywords{Conversational Question Answering, Adaptive Context Management, Natural Language Processing, Large Language Models.}
\end{abstract}

\section{Introduction}
Conversational Question Answering (ConvQA) systems, over the years, have become increasingly important for numerous applications, i.e., ranging from customer service to virtual assistants. These systems intend to provide accurate responses to a series of questions leveraging conversation history \cite{Qiu2021ReinforcedAnswering}. However, maintaining effective context in the ConvQA systems is challenging owing to existing large language models' limitations in handling extensive conversation histories\cite{Christmann2023ConversationalSources,Qu2019AttentiveAnswering}.
In ConvQA settings, a user begins with a question and the system responds using the provided context and the conversation history. Follow-up questions depend on previous conversation turns, thereby enabling the system to utilize conversation history for accurate responses \cite{Qu2019BERTAnswering}. Existing ConvQA approaches typically fall into two categories. The first category selects only the relevant context from the conversation history via either dynamic techniques \cite{Qiu2021ReinforcedAnswering,Zaib2021BERT-CoQAC:Context} or by prepending $K$ historical conversation turns \cite{KTurn1,Zhu2019SDNet:Answering}. These methods enhance a model's comprehension of the current question by selectively extracting relevant information from the conversation history and further reducing the context size by eliminating irrelevant information, thereby ensuring that the input remains within a model's token limit. However, methods falling under this category may omit critical information, in turn, reducing a model’s understanding and response accuracy. The second category employs the entire conversation history \cite{Ishii2022CanAnswering,Raposo2022QuestionAnswering} and is vulnerable to noise owing to too much irrelevant content. However, this is not a scalable approach as a model can only accommodate a fixed number of maximum tokens. Hence, as the conversation grows, a model would struggle and is left with no choice except to remove the older conversational turns from the conversation history. 

In order to address the aforementioned challenges, we envisaged an Adaptive Context Management (ACM) framework to optimize the use of conversation history by incorporating dynamic context management techniques -- summarization and entity extraction. Our envisaged approach prioritizes sending the maximum possible relevant context to a ConvQA model while effectively managing the context size to align with a model’s maximum token limit. As part of the ACM framework, we implemented three salient components, i.e., Context Manager (CM) Module, Summarization (SM) Module, and the Entity Extraction (EE) Module. The CM Module functions as a central hub that dynamically prioritizes and updates the conversation history for each question in a conversation thread. It ensures that the model accesses the optimal turns of the conversation history within its token limit when answering a user question. The SM Module summarizes the older parts of the conversation history within a summarization sliding window, thereby retaining indispensable information. The EE Module identifies and extracts key entities from the oldest parts of the conversational history once the summarization sliding window has been exceeded. This ensures that the crucial information is preserved without retaining the entire conversation history. The salient contributions of the paper-at-hand are as follows:

\begin{itemize}
    \item We envisaged the ACM framework which addresses the challenges of maintaining effective conversational context in ConvQA systems by dynamically managing conversation history. This ensures that the most relevant and recent conversation turns are used in a prioritized manner within a model’s maximum token limit, thereby enhancing response accuracy.

    \item We dynamically utilized the hard history selection technique in the ACM framework to ensure that the critical information is always accessible, thereby improving the robustness of the ConvQA models.
  
    \item We performed rigorous experiments on the coqa\_chat dataset, i.e, a subset of the CoQA dataset \cite{Reddy2019CoQA:Challenge}, for demonstrating the effectiveness of our envisaged ACM framework in terms of high accuracy and relevance of responses.
\end{itemize}

The rest of this particular paper is organized as follows. Section \ref{2} delineates the related work. Section \ref{3} outlines our envisaged ACM framework. Section \ref{4} describes the experimental setup, whereas, Section \ref{5} presents the experimental results and the comparative analysis. Finally, Section \ref{6} provides conclusions and discusses future work.

\section{Related Work}\label{2}

Significant advancements in ConvQA systems stem from the rapid progress of pre-trained large language models and large-scale datasets, i.e., QuAC and CoQA \cite{Zaib2023LearningAnswering}. These developments have led to state-of-the-art models and innovative strategies that enhance the accuracy and contextual relevance of ConvQA systems. This section reviews key research directions in ConvQA in terms of ConvQA question understanding and history selection approaches.

In ConvQA, Question Rewriting (QR) is a popular research direction aimed at addressing the challenges posed by incomplete or ambiguous questions. \cite{Ishii2022CanAnswering,Raposo2022QuestionAnswering,Kim2021LearnAnswering,Kaiser2024RobustGeneration} found that effective QR significantly improves a ConvQA system’s ability to retrieve information and generate accurate answers. This emphasized the need for advanced QR techniques. 
However, \cite{Zaib2023KeepingAnswering, Zaib2023LearningAnswering} argued that while QR can enhance a question's clarity, it often detaches the question from its conversation context by making the question context-independent. This can potentially lead to a loss of valuable data.

Accordingly, \cite{Zaib2023LearningAnswering} introduced DHS-ConvQA, a framework for replacing QR by utilizing context and question entities. This method pruned irrelevant conversation turns and re-ranked the rest of the data based on relevance. This approach outperformed QR in ConvQA performance on the CANARD and QuAC datasets. Similarly, \cite{Zaib2023KeepingAnswering} proposed CONVSR, a framework that captured intermediate structured representations to resolve dependencies in ConvQA systems. This approach enhanced the ConvQA system's performance on QuAC and CANARD datasets through dynamic history selection and context-question entities.

A key challenge in ConvQA is effectively incorporating conversation history to accurately interpret the current question. Understanding the current question often requires context from previous interactions. This makes it essential to manage and utilize the conversation history effectively. Various strategies have been developed to address this challenge. One common approach is to prepend $K$ historical conversation turns to the current question \cite{KTurn1,Zhu2019SDNet:Answering}. This method ensures that the model can access relevant previous interactions, thereby improving its ability to generate accurate responses and enhancing performance on tasks involving conversational context. In the context of prepending $K$ historical conversation turns, some authors \cite{Kim2021LearnAnswering,Qu2019BERTAnswering,semantic1}
utilized only the immediate conversation turn based on the intuition that an immediate conversation turn was typically more relevant to the current question. Dynamic history selection \cite{Zaib2023KeepingAnswering, Zaib2023LearningAnswering,Zaib2021BERT-CoQAC:Context} was another crucial approach and can be categorized into hard and soft history selection techniques. Hard history selection involves choosing a subset of question-relevant conversation turns and effectively filtering out irrelevant conversation turns. Conversely, soft history selection involves assigning different weights to various parts of the conversation history, thereby allowing the system to consider entire historical conversation turns but with varying degrees of importance. For instance, a reward-based reinforcement learning approach was employed in \cite{Qiu2021ReinforcedAnswering} to refine the selection of conversation history dynamically. This method used a reward mechanism to iteratively improve the selection of historical conversation turns. It ensured that the most relevant information was retained for answering the current question. In our research, we leveraged hard history selection technique to manage conversation history dynamically. By employing a dynamic history selection method, we aim to enhance the model’s ability to retain relevant historical context while filtering out noise, thereby improving the overall performance of our ConvQA system.

\section{The ACM Framework}\label{3}

\begin{figure}[!t]
    \centering
    \includegraphics[width=0.8\textwidth]{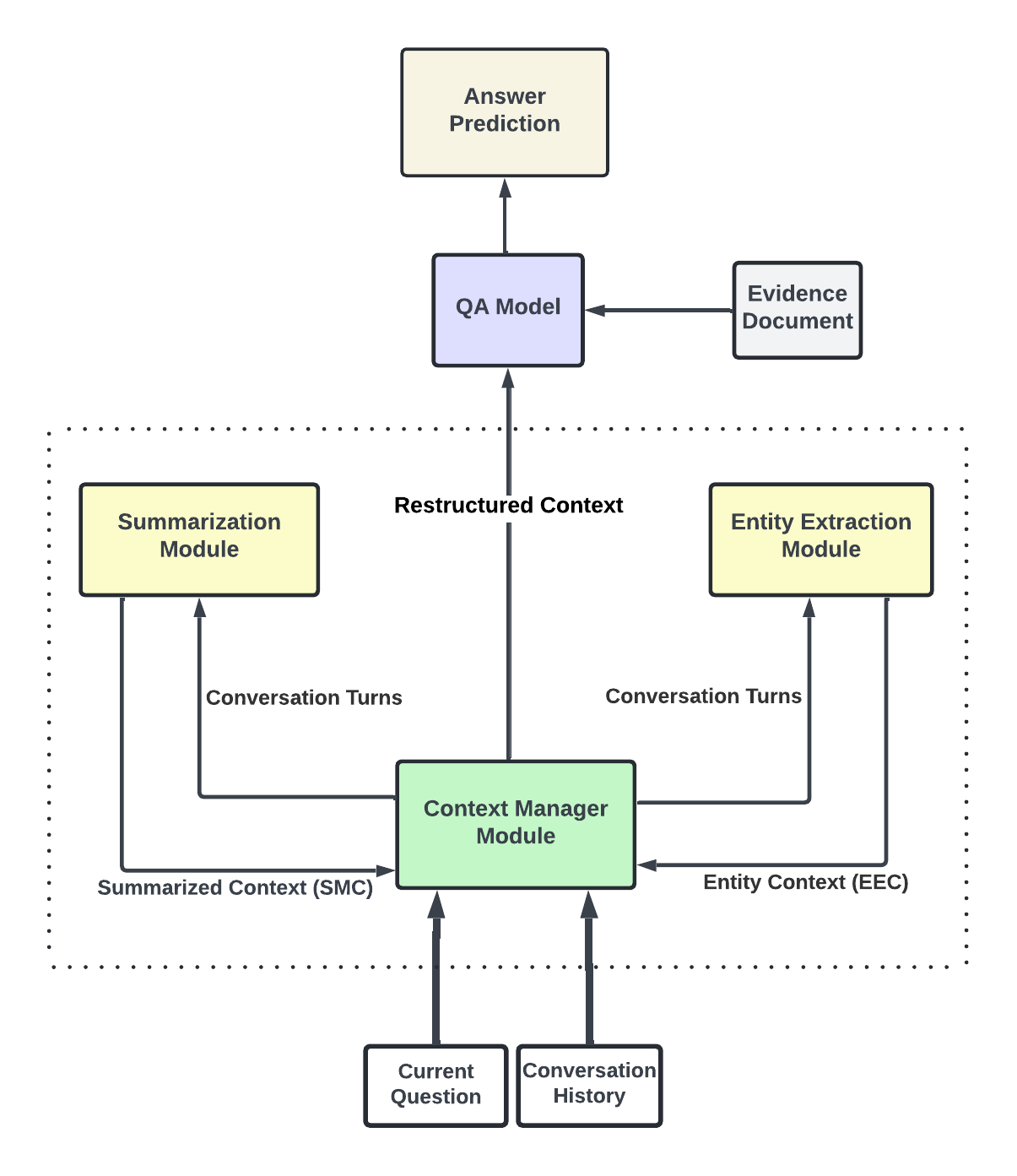} % Adjust the width as needed
    \caption{System architecture for the envisaged ACM framework. The CM Module dynamically maintains and updates the context, integrating inputs from the SM Module and the EE Module, to generate accurate responses using the QA Model.}
    \label{Fig.1:acm_architecture}
\end{figure}

The ACM framework optimizes context management in ConvQA systems by prioritizing pertinent conversation history, thereby enhancing the system's ability to generate accurate responses. This framework utilizes the most relevant and recent conversation turns without overwhelming the ConvQA model with too much irrelevant information. As depicted in Fig. \ref{Fig.1:acm_architecture}, the ACM framework comprises three main modules -- CM Module, SM Module, and EE Module. Each of these modules plays a crucial role in dynamically managing and optimizing the context provided to the ConvQA model. As depicted in Fig. \ref{fig:context_management}, the CM Module prioritizes sending the maximum possible context to the ConvQA model. This approach prioritizes recent conversation turns and preserves them in full detail, whereas, older conversation turns are summarized and key entities are extracted from the oldest conversation turns. This ensures the most recent and relevant context is fully available to the ConvQA model. The ACM framework dynamically adjusts the context size to align with a model's token limit, thereby optimizing both the quantity and quality of the conversation history. This balance enhances the ConvQA model's ability to generate accurate and contextually appropriate responses.

\subsection{\textbf{Context Manager (CM) Module}}

\begin{figure}[!t]
    \centering
    \includegraphics[width=\textwidth]{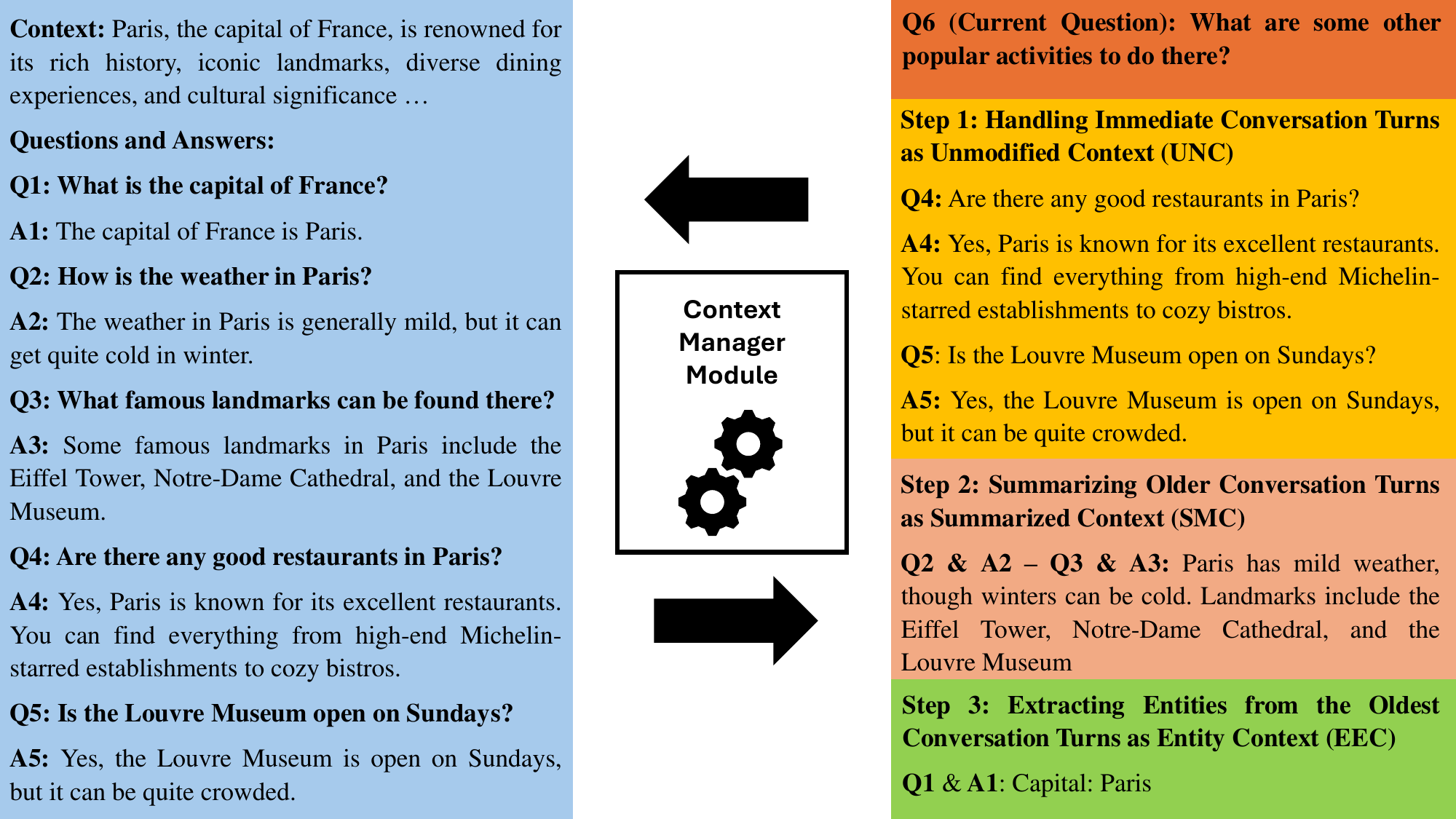} % Set the width to the full text width
    \caption{Depicting context management process in the CM Module -- the diagram illustrates the handling of the recent conversation turns, summarization of the older conversation turns, and extraction of the key entities from the oldest conversation turns.}
    \label{fig:context_management}
\end{figure}

The CM Module, as portrayed in Fig. \ref{fig:context_management}, is an indispensable constituent of the ACM framework. It dynamically maintains and updates the context for each question in a conversation thread. The key aim of the CM Module is to ensure that a model accesses the optimal turns of the conversation history within its token limit. The CM Module thus functions as a central hub. It is involved in (a) \textit{preprocessing of the data --} preparing and structuring the conversational data for an effective use, (b) \textit{training ConvQA models --} facilitating the training of the models in a bid to handle the conversational data efficiently, and (c) \textit{managing the inference --} overseeing the real-time processing of the conversational inputs to generate accurate responses. The CM Module integrates two main components, i.e., SM Module and the EE Module. These components work together to manage the conversation history in a dynamic manner.

\subsection{Summarization (SM) Module}\

Text Summarization is one of the most important Natural Language Processing (NLP) concepts \cite{Chen2020Multi-ViewSummarization, Jin2024AMethods}. Summarizing conversation history is crucial for maintaining the efficiency and effectiveness of a ConvQA system. It ensures that essential information is retained while keeping the context size within a manageable limit, thereby optimizing the model's performance. The SM Module summarizes the older parts of the conversation history. It performs this action when the unmodified context size exceeds the model's token limit. This ensures that essential information is retained while managing the context size. We conducted a series of experiments as shown in \cref{fig:bart_conversation,fig:bart_large,fig:pegasus_large,fig:t5_large} to determine the optimal token length for summarizing conversation history. The key benefits of the SM Module include:

\begin{itemize}
    \item \textbf{Improving Model Efficiency:} The SM Module significantly enhances the efficiency of the ConvQA system by reducing the context size without sacrificing essential information. By summarizing older conversation turns, the ACM framework ensures that the ConvQA model processes a manageable amount of data, thereby optimizing computational resources and reducing processing time.

    \item \textbf{Improving Response Accuracy:} By retaining the most critical information from the conversation history, the SM Module ensures that the ConvQA model has access to all necessary context to generate accurate and relevant responses. This improves the overall accuracy of the system, as the model can focus on the most pertinent details.
    
    \item \textbf{Improving Scalability:} The ability to summarize older conversation turns allows the ConvQA system to handle longer and more complex conversation histories. This scalability is crucial for real-world applications where conversations can span numerous interactions, thereby ensuring that the system remains effective even as the amount of data grows.
    
\end{itemize}
	
\subsection{\textbf{Entity Extraction (EE) Module}} 

Once the summary sliding window has been exceeded, we need to manage older parts of the conversation history. In order to handle this overflow of the conversation turns, we employ an EE Module. This module extracts key entities from the oldest conversation turns for retaining crucial information without preserving the entire text. The goal is to capture essential details, i.e., names, dates, and locations, ensuring important context is maintained within a model's token limit. 

The EE Module operates in three main steps. During the \textit{entity identification phase}, the model scans the oldest conversation turns and identifies key entities using Named Entity Recognition (NER) techniques. These techniques leverage advanced NLP algorithms to detect and classify entities into predefined categories, i.e., people, organizations, dates, and locations \cite{Reshmi2018EnhancingIntegration}. This process ensures that significant pieces of information are accurately recognized from the textual data. In the \textit{entity extraction phase}, the identified entities are extracted from the text and stored in a structured format. This step is crucial for maintaining the integrity of the information while reducing the overall text length. By isolating entities, the model can discard surrounding text that does not significantly contribute to the context, thereby saving valuable token space. Finally, during the \textit{context integration phase}, the extracted entities are integrated into the current context and replace the full text of the oldest conversation turns. This integration ensures that the critical information from the older parts of the conversation history is retained in a concise form. Additionally, this approach of using entity extraction provides several benefits. It enhances the model's efficiency by focusing on essential information and reducing noise from irrelevant text. Furthermore, it ensures that the system remains scalable and responsive, even as the conversation history grows. By systematically extracting and integrating key entities, the EE Module supports the overarching goal of maintaining a high-quality context that enables the ConvQA system to generate accurate and contextually appropriate responses.

\subsection{Dynamic Context Window Adjustment (DCWA)}

\begin{algorithm}[!t]
\caption{Dynamic Context Window Adjustment}
\begin{algorithmic}[1]
\REQUIRE: Context: $C_n$, Base Passage: \text{$Context_B$}, Summarized Context: $SMC$, Entity Context: $EEC$, Unmodified Context:
$UNC$, Model Token Size: $MS_{\max}$, %Summary Size: $C_{\text{sum}}$,
Summary Limit: $SM_{\text{limit}}$, %Entity size: $C_{\text{ent}}$,
%Entity Limit: $EE_{\text{limit}}$,
Turns: $T$,
Minimum Unmodified Context \%: $Threshold$, Token Size of the Given Text: $TokenCount$.

\FOR {each question $Q_n$ in Conversation}

    \IF{$Q_n$ is the first question}
        \STATE $C_n = \text{$Context_B$} + Q_1$
    \ELSE
        \STATE $C_n = \text{$Context_B$} + Q_n +UNC( \sum_{j=1}^{n-1} T_{j})$
    \ENDIF
    \IF{$TokenCount$($C_n$) $>$ $MS_{\max}$}
        \STATE $C_n = \text{$Context_B$} + Q_n +SMC( \sum_{j=1}^{m-1} T_{j}) + UNC(\sum_{i=m}^{n-1} T_{i})$
    \ENDIF
    \IF{$TokenCount(SMC( \sum_{j=1}^{m-1} T_{j})) $ > $SM_{\text{limit}}$}
        \WHILE{$TokenCount(UNC(\sum_{i=m}^{n-1} T_{i})) $>$ ($Threshold$ \times MS_{\max}$)}
            \STATE $oldest\_turn\_size = TokenCount$($UNC(T_m)$)
            \IF{($TokenCount(UNC(\sum_{i=m}^{n-1} T_{i})) - oldest\_turn\_size) \geq (Threshold \times MS_{\max}$)}
                \STATE $EEC = EEC + oldest\_turn\_size$
                \STATE $UNC = UNC - oldest\_turn\_size$
                \STATE $C_n = \text{$Context_B$} + Q_n +EEC( \sum_{j=1}^{m-1} T_{j}) + SMC( \sum_{k=m}^{p-1} T_{k}) + UNC( \sum_{i=p}^{n-1} T_{i})$
            \ELSE
                \STATE \textbf{break}
            \ENDIF
        \ENDWHILE
    \ENDIF
\ENDFOR   
\end{algorithmic}
\end{algorithm}

The CM Module's primary function is to ensure that a model can access the optimal level of historical conversation turns within its maximum token limit. To achieve this, the CM Module employs a technique known as Dynamic Context Window Adjustment (DCWA), which optimizes token allocation across different parts of the context.
Initially, the entire context window is allocated for Unmodified Context ($UNC$). $UNC$ includes the \text{$Context_B$} which refers to the base passage utilized for answering questions
in the ConvQA model and the recent conversation turns without any modifications. The CM Module processes each question in the conversation as follows:

\subsubsection{Initial Question:}
 For the first question ($Q_1$), the model processes the \text{$Context_B$} combined with the current question as depicted in Equation \ref{EQ1}.
      
\begin{equation}\label{EQ1}
C_n = \text{$Context_B$} + Q_1
\end{equation}

\subsubsection{Subsequent Questions:}
 For each subsequent question, the input includes the \text{$Context_B$}, the current question ($Q_n$), and the previous ($n-1$) conversation turns as $UNC$, as depicted in Equation \ref{EQ2}.
    
\begin{equation}\label{EQ2}
C_n = \text{$Context_B$} + Q_n +UNC( \sum_{j=1}^{n-1} T_{j})
\end{equation}

The CM Module continues this process until the context($C_n$) exceeds the model's maximum token limit ($MS_{\max}$).
When the $C_n$ exceeds the $MS_{\max}$, the CM Module initiates a sliding window for summarization, thereby summarizing older conversation turns in the $UNC$ within the sliding window. It subsequently incorporates the latest conversation turns as $UNC$, along with the Summarized Context ($SMC$), into the ConvQA model as depicted in Equation \ref{EQ3}.

\begin{equation}\label{EQ3}
C_n = \text{$Context_B$} + Q_n +SMC( \sum_{j=1}^{m-1} T_{j}) + UNC(\sum_{i=m}^{n-1} T_{i})
\end{equation}

As the conversation grows, if the $SMC$ exceeds the $SM_{limit}$, EE Module can borrow tokens from the oldest turn of the $UNC$, thereby creating a sliding window for entity context ($EEC$). The Entity window captures the critical information as keywords from the oldest conversation turns in the $SMC$ as depicted in Equation \ref{EQ4}.

\begin{equation}\label{EQ4}
C_n = \text{$Context_B$} + Q_n +EEC( \sum_{j=1}^{m-1} T_{j}) + SMC( \sum_{k=m}^{p-1} T_{k}) + UNC( \sum_{i=p}^{n-1} T_{i})
\end{equation}

This adjustment continues until the $UNC$ reaches the \(Threshold\)$X$\(MS_{max}\). Following these adjustments, the $UNC$ occupies a reduced yet still significant portion of the overall \(MS_{max}\). Concurrently, the $SMC$ and $EEC$ are allocated space to handle critical information from older conversational turns efficiently. By incorporating this dynamic adjustment technique, the CM Module ensures that the most recent and pertinent information is readily accessible to the ConvQA model while effectively managing the $C_n$.

\section{Experimental Setup}\label{4}

\begin{figure}[h!]
    \centering
    \begin{minipage}[t]{0.48\textwidth} % Increased size from 0.45 to 0.48
        \centering
        \includegraphics[width=\textwidth]{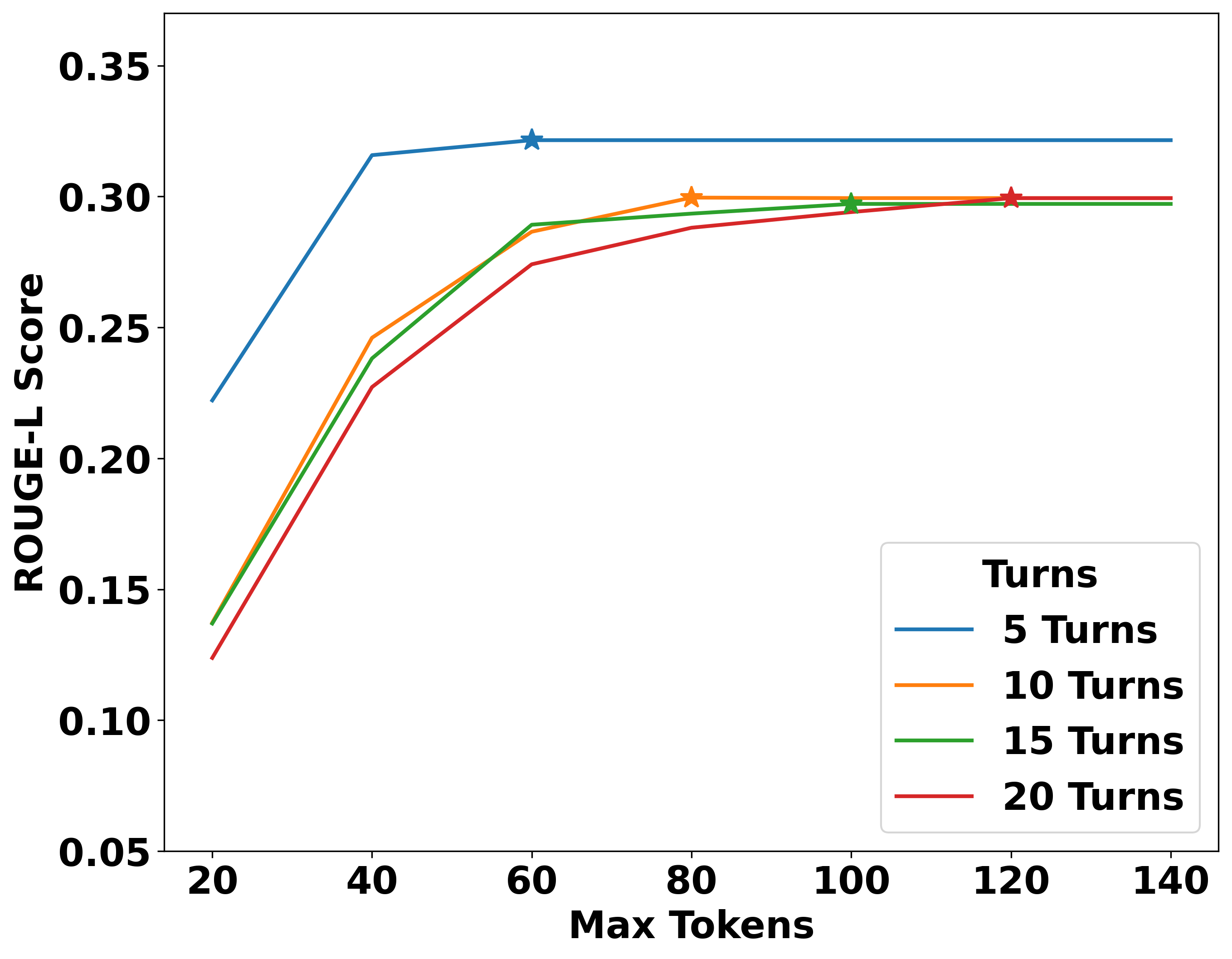}\label{Fig3}
        \caption{ROUGE-L Score vs. Max Tokens (Bart Conversation Summary).}
        \label{fig:bart_conversation}
    \end{minipage}
    \hfill
    \begin{minipage}[t]{0.48\textwidth} % Increased size from 0.45 to 0.48
        \centering
        \includegraphics[width=\textwidth]{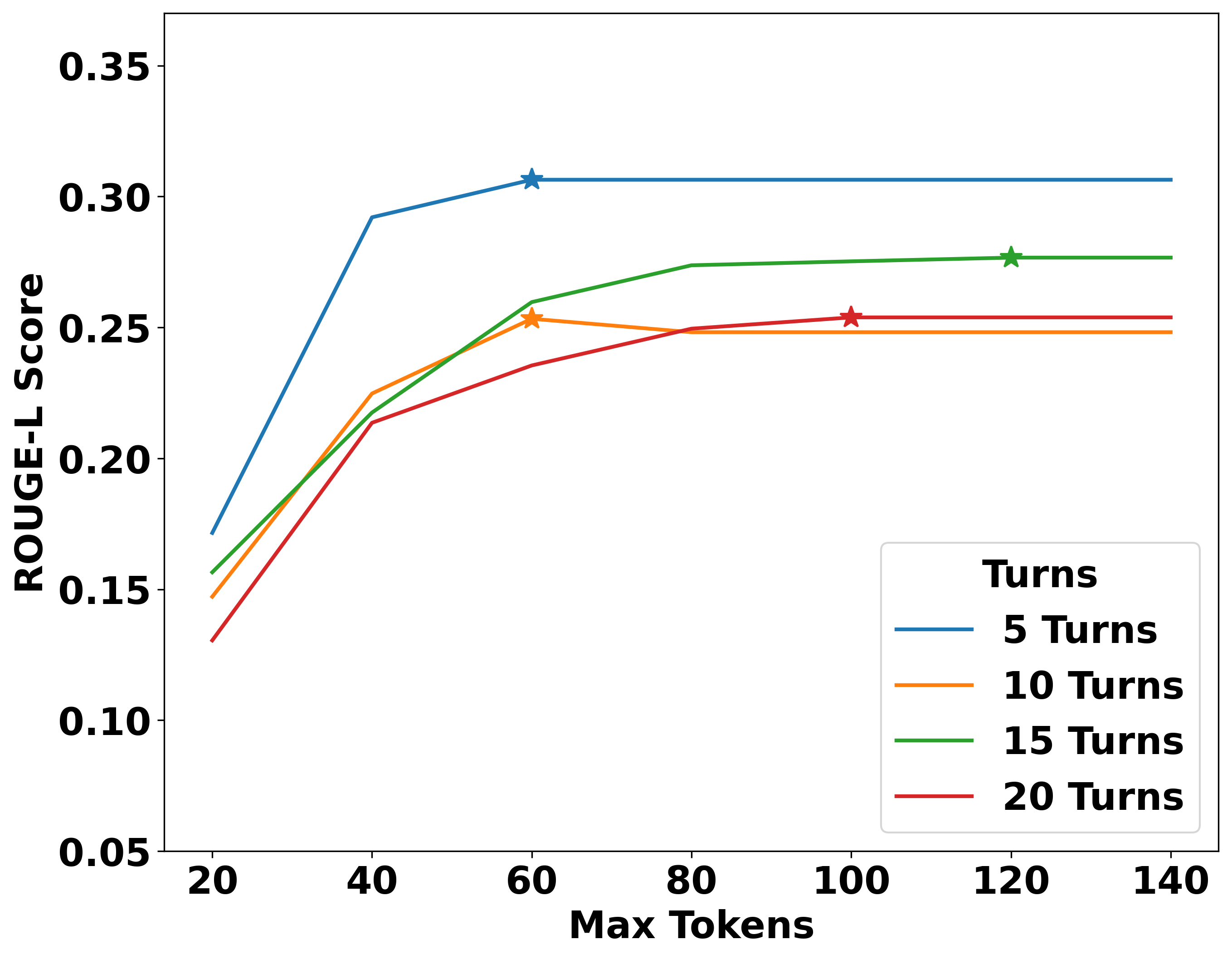}\label{Fig4}
        \caption{ROUGE-L Score vs. Max Tokens (Bart Large CNN).}
        \label{fig:bart_large}
    \end{minipage}

    \vspace{0.5cm}

    \begin{minipage}[t]{0.48\textwidth} % Increased size from 0.45 to 0.48
        \centering
        \includegraphics[width=\textwidth]{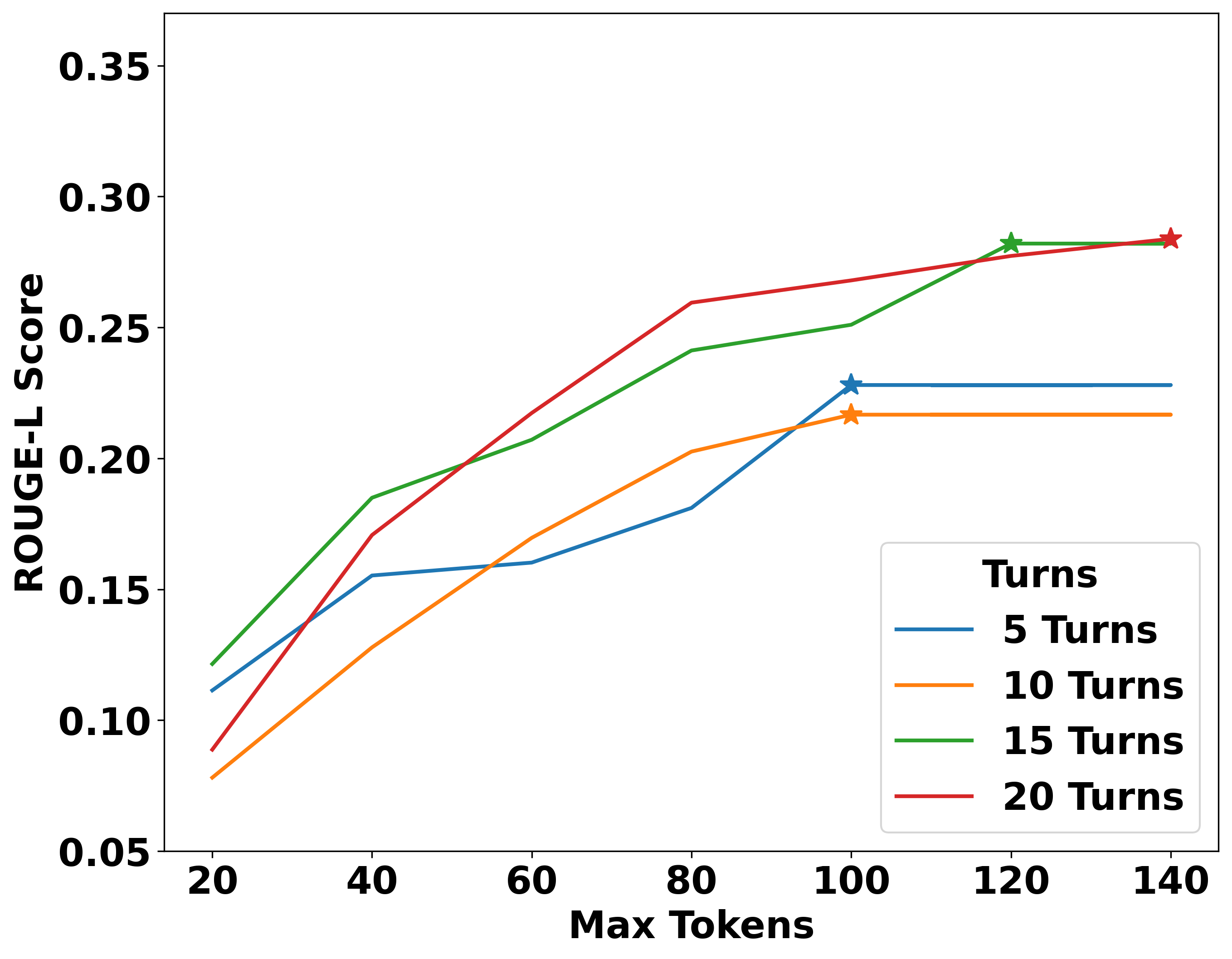}\label{Fig5}
        \caption{ROUGE-L Score vs. Max Tokens (Pegasus Large).}
        \label{fig:pegasus_large}
    \end{minipage}
    \hfill
    \begin{minipage}[t]{0.48\textwidth} % Increased size from 0.45 to 0.48
        \centering
        \includegraphics[width=\textwidth]{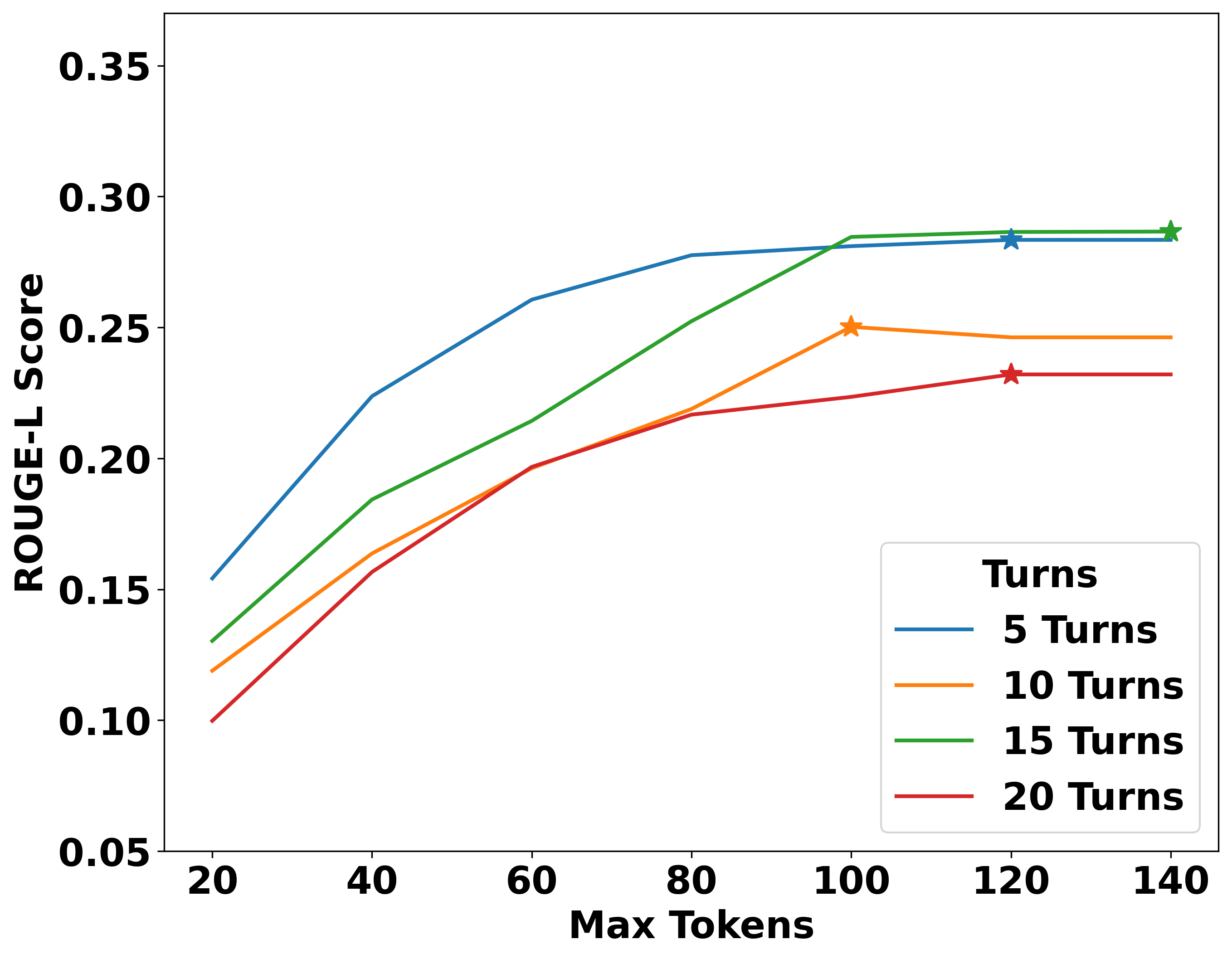}\label{Fig3}
        \caption{ROUGE-L Score vs. Max Tokens (T5 Large).}
        \label{fig:t5_large}
    \end{minipage}
\end{figure}

We conducted a series of experiments to identify the optimal number of tokens required for the effective summarization of older parts of the conversation history. We generated summaries using a range of large language models and evaluated their efficacy based on the ROUGE-L score \cite{Bhandari2020Re-evaluatingSummarization, El-Kassas2020EdgeSumm:Summarization,ROUGE-ChinYew}. 
The selected state-of-the-art large language models include -- \texttt{BART Conversation Summary} \cite{Lewis2020BART:Comprehension}, \texttt{BART Large CNN} \cite{Bajaj2021LongModels}, \texttt{Pegasus Large} \cite{Zhang2020PEGASUS:Summarization,G2024ComparativeSummarization}, and \texttt{T5 Large} \cite{G2024ComparativeSummarization,Jadeja2022ComparativeDataset}. For each large language model, we generated summaries using different numbers of conversation turns (5, 10, 15, and 20) from the coqa\_chat dataset. The generated summaries were then compared to a golden summary created via \texttt{gemini-1.5-pro} large language model \cite{reid2024gemini}. \cref{fig:bart_conversation,fig:bart_large,fig:pegasus_large,fig:t5_large}
portray the correlation between the maximum tokens and the ROUGE-L score for each model. As can be observed, since the \texttt{BART Conversation Summary} model consistently demonstrated higher ROUGE-L score across different conversation turns, we employed the same model as the summarization model for our envisaged ACM framework. 

Specifically, the \texttt{BART Conversation Summary} model requires 60, 80, 100, 120 tokens for 5, 10, 15, 20 conversation history turns to generate accurate summaries.
Based on this analysis, we opted for 120 tokens as the maximum tokens for the summarization sliding window. This decision was made on the observation that 120 tokens effectively capture sufficient context for longer conversation turns (up to 20 turns), thereby ensuring the retention of a high ROUGE-L score in our evaluations.
Additionally, this sliding window can accommodate variations in conversation length and complexity, thereby providing a flexible yet effective summarization.
The allocated tokens can be varied based on the selected model and the dataset. This flexibility is crucial as different models and datasets might exhibit different characteristics and requirements for effective summarization. By opting to the tune of 120 tokens, we ensure that our summarization process remains robust across diverse conversational contexts, thereby minimizing the risk of losing critical information due to token constraints. Moreover, it is evident from \cref{fig:bart_conversation,fig:bart_large,fig:pegasus_large,fig:t5_large} that shorter conversation turns (5 -- 10 turns) generally required fewer tokens to achieve a high ROUGE-L score. Conversely, longer conversation turns (15 -- 20 turns) benefited from a larger token limit, thereby allowing the model to capture more context and produce more coherent summaries.

\subsection{Dataset}

We conducted experiments using the coqa\_chat dataset, which is an amended version of CoQA dataset \cite{Reddy2019CoQA:Challenge}, designed to evaluate large language models' ability to handle ConvQA. The coqa\_chat dataset contains over 117k question-answer pairs. These question-answer pairs have been adjusted to be more conversational, focusing on providing contextually relevant information along with the answers. The coqa\_chat dataset enhances conversational relevance and is, therefore, making it ideal for training large language models for in-context and document question-answering.

\subsection{Metrics}

To evaluate the performance of our envisaged ACM framework, we used four key metrics, i.e., F1 score \cite{Zaib2023KeepingAnswering}, ROUGE-1 score \cite{ROUGE-ChinYew}, ROUGE-L score \cite{ROUGE-ChinYew}, and BLEU score  \cite{ROUGE-ChinYew}. The F1 score measures the balance between precision and recall manifesting a comprehensive assessment of accuracy. ROUGE-1 assesses the overlap of unigrams (single words) between the generated text and the reference suggesting a direct measure of content preservation. ROUGE-L evaluates the longest common sub-sequence and thus captures the fluency and coherence of the generated text in relation to the reference. ROUGE-L was used to assess both the ACM framework and the summarization model. The BLEU score assesses the accuracy and fluency of machine-generated text against human-generated references. Together, these metrics offer a robust evaluation of the systems' performance in generating accurate and contextually appropriate responses.

\subsection{Pipeline Approach}

To evaluate the accuracy of the ACM framework, we opted for six state-of-the-art large language models, i.e, \texttt{RoBERTa}, \texttt{BERT-Large}, \texttt{mDeBERTa-v3}, \texttt{Llama-2-7b}, \texttt{Mistral-7B}, and \texttt{phi-2}. These models were chosen for their strong performance on question-answering tasks and their ability to handle diverse and complex queries. To compare the performance of our envisaged ACM framework, we implemented a pipeline approach that utilizes the immediate conversation turn approach for focusing on the most recent conversation turn to generate responses. This approach was chosen to provide a direct comparison with our ACM framework in a bid to highlight the improvements and advantages of our framework in maintaining conversational context and generating accurate, contextually appropriate responses.

\section{Experimental Results and Comparative Analysis}\label{5}
 % Add this line in your preamble if not already included
\begin{table}[!t]
\centering
\scriptsize % Reduce font size further
\setlength{\tabcolsep}{3pt} % Adjust the space between columns
\renewcommand{\arraystretch}{1.2} % Adjust the row height
\begin{tabular}{>{\centering\arraybackslash}p{1.8cm}|>{\centering\arraybackslash}p{2.2cm}|>{\centering\arraybackslash}p{1.5cm}|>{\centering\arraybackslash}p{1.5cm}|>{\centering\arraybackslash}p{1.5cm}|>{\centering\arraybackslash}p{1.5cm}}
\hline
\textbf{Model} & \textbf{Evaluation Method} & \textbf{F1 Score} & \textbf{ROUGE-L Score} & \textbf{ROUGE-1 Score} & \textbf{BLEU Score} \\
\hline
\multirow{2}{*}{phi-2} & Pipeline & 58.57 & 55.91 & 60.08 & 46.83 \\
 & \textbf{ACM Approach} & \textbf{69.35 \textcolor{blue}{(+10.78)}} & \textbf{66.34 \textcolor{blue}{(+10.43)}} & \textbf{72.98 \textcolor{blue}{(+12.90)}} & \textbf{55.23 \textcolor{blue}{(+8.40)}} \\
\hline
\multirow{2}{*}{Mistral-7B} & Pipeline & 56.31 & 54.44 & 58.13 & 45.13 \\
 & \textbf{ACM Approach} & \textbf{66.71 \textcolor{blue}{(+10.40)}} & \textbf{63.12 \textcolor{blue}{(+8.68)}} & \textbf{68.24 \textcolor{blue}{(+10.11)}} & \textbf{52.34 \textcolor{blue}{(+7.21)}} \\
\hline
\multirow{2}{*}{Llama-2-7b} & Pipeline & 51.11 & 50.45 & 54.79 & 42.48 \\
 & \textbf{ACM Approach} & \textbf{60.21 \textcolor{blue}{(+9.10)}} & \textbf{57.73 \textcolor{blue}{(+7.28)}} & \textbf{63.87 \textcolor{blue}{(+9.08)}} & \textbf{48.79 \textcolor{blue}{(+6.31)}} \\
\hline
\multirow{2}{*}{mDeBERTa-v3} & Pipeline & 48.41 & 46.77 & 49.91 & 35.09 \\
 & \textbf{ACM Approach} & \textbf{54.46 \textcolor{blue}{(+6.05)}} & \textbf{51.89 \textcolor{blue}{(+5.12)}} & \textbf{56.29 \textcolor{blue}{(+6.38)}} & \textbf{40.51 \textcolor{blue}{(+5.42)}} \\
\hline
\multirow{2}{*}{RoBERTa} & Pipeline & 47.22 & 45.61 & 48.88 & 34.66 \\
 & \textbf{ACM Approach} & \textbf{53.12 \textcolor{blue}{(+5.90)}} & \textbf{51.02 \textcolor{blue}{(+5.41)}} & \textbf{54.96 \textcolor{blue}{(+6.08)}} & \textbf{39.55 \textcolor{blue}{(+4.89)}} \\
\hline
\multirow{2}{*}{BERT-Large} & Pipeline & 46.78 & 43.26 & 48.22 & 32.89 \\
 & \textbf{ACM Approach} & \textbf{52.51 \textcolor{blue}{(+5.73)}} & \textbf{49.54 \textcolor{blue}{(+6.28)}} & \textbf{54.05 \textcolor{blue}{(+5.83)}} & \textbf{38.07 \textcolor{blue}{(+5.18)}} \\
\hline
\end{tabular}
\vspace{0.5em} % Add space between table and caption

\caption{Performance comparison of the envisaged ACM framework vis-\`a-vis the pipeline approach in the context of state-of-the-art large language models.} 
\label{table:1}
\end{table}

In this particular section, we present a comprehensive analysis pertinent to the performance of our envisaged ACM framework vis-\`a-vis the pipeline approach in the context of six state-of-the-art large language models. The evaluation metrics include F1 score, ROUGE-L score, ROUGE-1 score, and BLEU score. We trained and subsequently tested all the six large language models under both the ACM framework and pipeline approach using a 10\% sample from the coqa\_chat dataset, i.e., which is well-regarded in the research community for its follow-up questions. When performing these experimental results, we fixed the minimum $Threshold$ for \(UNC\) to 75\%. Furthermore, we utilized \texttt{en\_core\_web\_sm} model from spaCy, a popular NLP library, to extract key entities from the oldest conversation turns, thereby ensuring that crucial information was retained while managing the token limit effectively. As depicted in the Table \ref{table:1}, our envisaged ACM framework considerably outperformed the pipeline approach for all the evaluation metrics. This thus demonstrates the effectiveness of the ACM framework in handling the conversation history. The key findings of the experimental results are delineated as follows:

\begin{itemize}
    \item \textbf{F1 Score:} Across all of the six large language models, the F1 score improved significantly in the case of our envisaged ACM framework. This, in essence, indicates better precision and recall, thereby demonstrating the ACM framework's ability to ascertain more accurate and relevant answers.
    \item \textbf{ROUGE-L Score:} Similar to the F1 score, the ROUGE-L score improved considerably for the ACM framework. This improved ROUGE-L score suggests that the ACM framework produces answers that are more coherent and contextually aligned with the referenced answers, thereby effectively capturing the flow and the structure of the conversation.
    \item \textbf{ROUGE-1 Score:} The ROUGE-1 score, which evaluates the overlap of the unigrams (single words) amongst the generated responses and the referenced answers, showed considerable improvement for the ACM framework too. This indicates a precise match of the individual words hence reflecting the ACM framework's strength in including relevant keywords for enhancing the overall relevance and accuracy.
    \item \textbf{BLEU Score:} The BLEU score evaluates both fluency and the grammatical correctness of the responses, and is higher for the ACM framework indicating that the generated answers were accurate and linguistically sound.
\end{itemize}

The significant improvement in the evaluation metrics manifests the efficacy of the envisaged ACM framework in managing the conversation history. By dynamically adjusting the context size together with employing summarization and entity extraction techniques, our framework ensures that the most relevant information is preserved and utilized, thereby leading to more accurate and coherent responses in the ConvQA systems even if a particular model's maximum token limit has been exceeded.

\section{Conclusion and Future Work}\label{6}

In this paper, we introduced the ACM framework for addressing the challenges pertinent to maintaining effective context in the ConvQA systems. By integrating the CM, SM, and EE Modules, the ACM framework dynamically prioritizes and manages the conversation history. This guarantees that the most relevant and recent information is utilized within a large language model’s maximum token limit, thereby resulting in improved accuracy and contextual appropriateness of the responses. Despite the inherent limitations of the large language model’s contextual capacity, the envisaged framework effectively maximizes usable context by dynamically adjusting the input hence demonstrating its scalability. Our experimental results, validated on the coqa\_chat dataset, demonstrate the significant performance gains achieved by the ACM framework vis-\`a-vis the pipeline approach in the context of six large language models. These findings highlight the potential of dynamic context management techniques to enhance the robustness and effectiveness of the ConvQA systems.

Future research will focus on refining the components of the ACM framework to further enhance its efficiency and effectiveness. One promising direction is the integration of more advanced summarization and entity recognition techniques so as to improve the precision of the context management. Expanding the evaluation to include a broad range of datasets and conversational scenarios will also be crucial for validating the generalizability of our envisaged ACM framework. Finally, investigating the application of the ACM framework in real-world ConvQA applications, i.e., customer service and virtual assistants, will further help to assess its practical utility and impact.

\bibliographystyle{IEEEtran}
\bibliography{ReferencesTechnical}

\end{document}